\documentclass[letterpaper, 10 pt, conference]{ieeeconf}  % Comment this line out if you need a4paper

\IEEEoverridecommandlockouts                              % This command is only needed if 
\usepackage{amsmath} % assumes amsmath package installed
\usepackage{amssymb}  % assumes amsmath package installed
\usepackage{graphicx}
\usepackage{multirow}
\usepackage{booktabs}
\usepackage{float}
\usepackage{tikz}
\usepackage{textcomp}
\usepackage{hyperref}
\usepackage{lipsum}
\graphicspath{ {Figures/} } 

\newcommand\copyrighttext{%
  \footnotesize \textcopyright 2021 IEEE. Personal use of this material is permitted.
  Permission from IEEE must be obtained for all other uses, in any current or future
  media, including reprinting/republishing this material for advertising or promotional
  purposes, creating new collective works, for resale or redistribution to servers or
  lists, or reuse of any copyrighted component of this work in other works.
  DOI: \href{<http://tex.stackexchange.com>}{<DOI No.>}}
\newcommand\copyrightnotice{%
\begin{tikzpicture}[remember picture,overlay]
\node[anchor=south,yshift=10pt] at (current page.south) {\fbox{\parbox{\dimexpr\textwidth-\fboxsep-\fboxrule\relax}{\copyrighttext}}};
\end{tikzpicture}%
}

\title{\LARGE \bf
Gait-based Human Identification through Minimum Gait-phases and Sensors *
}

\author{Muhammad Zeeshan Arshad, Dawoon Jung, Mina Park, Kyung-Ryoul Mun, and Jinwook Kim% <-this % stops a space
\thanks{This research was supported by the Korea Institute of Science and Technology Institutional Program (Project No. 2E31051) and in part by the High-Tech Based National Athletic Performance Improvement (Winter) Project from the Korea Sports Promotion Foundation.}%

\thanks{$^{1}$Muhammad Zeeshan Arshad, Dawoon Jung, Mina Park, Kyung-Ryoul Mun, and Jinwook Kim are with the Center for Artificial Intelligence, KIST, Seoul, Republic of Korea.
        {\tt\small jwkim@imrc.kist.re.kr}}%
}

\begin{document}

\maketitle
\copyrightnotice
\thispagestyle{empty}
\pagestyle{empty}

%%%%%%%%%%%%%%%%%%%%%%%%%%%%%%%%%%%%%%%%%%%%%%%%%%%%%%%%%%%%%%%%%%%%%%%%%%%%%%%%
\begin{abstract}

The incredible pace at which the world's elderly population is growing will put severe burdens on current healthcare systems and resources. To alleviate this concern the health care systems must rely on the transformation of eldercare and old homes to use Ambient Assisted Living (AAL). Human identification is one of the most common and critical tasks for condition monitoring, human-machine interaction, and providing assistive services in such environments. Recently, human gait has gained new attention as a biometric for identification to achieve contactless identification from a distance robust to physical appearances. However, an important aspect of gait identification through wearables and image-based systems alike is accurate identification when limited information is available for example, when only a fraction of the whole gait cycle or only a part of the subject's body is visible. In this paper, we present a gait identification technique based on temporal and descriptive statistic parameters of different gait phases as the features and we investigate the performance of using only single gait phases for the identification task using a minimum number of sensors. Gait data were collected from 60 individuals through pelvis and foot sensors. Six different machine learning algorithms were used for identification. It was shown that it is possible to achieve high accuracy of over 95.5\% by monitoring a single phase of the whole gait cycle through only a single sensor. It was also shown that the proposed methodology could be used to achieve 100\% identification accuracy when the whole gait cycle was monitored through pelvis and foot sensors combined. The ANN was found to be more robust to less number of data features compared to SVM and was concluded as the best machine algorithm for the purpose.        

\end{abstract}

%%%%%%%%%%%%%%%%%%%%%%%%%%%%%%%%%%%%%%%%%%%%%%%%%%%%%%%%%%%%%%%%%%%%%%%%%%%%%%%%
\section{INTRODUCTION}
The significant advancements in medical science and technology in the last few decades have caused an unprecedented transformation in global demographics. Life expectancy has increased at a rapid scale, so much so that according to a recent United Nation’s (UN) report, the world elderly population is expected to grow more than double its current size by 2050 \cite{nations2020world}. On the other hand, fewer and fewer people in the younger age group will be available to provide for the increasing health care demands of the elderly. This calls for revolutionizing the elderly homes and health care service centers through the realization of Ambient Assisted Living (AAL) systems \cite{dohr2010internet}. The recent rapid growth of 5G technologies, wireless sensor networks, and wearable sensors together with tremendous advancements in the Machine Learning (ML) domain, has opened new possibilities for complete automation of services through continuous and unobtrusive tracking, condition monitoring, diagnostics, and risk prevention \cite{ahad20195g}.

\begin{figure*}[thpb]
\centering
  \includegraphics[width=\linewidth]{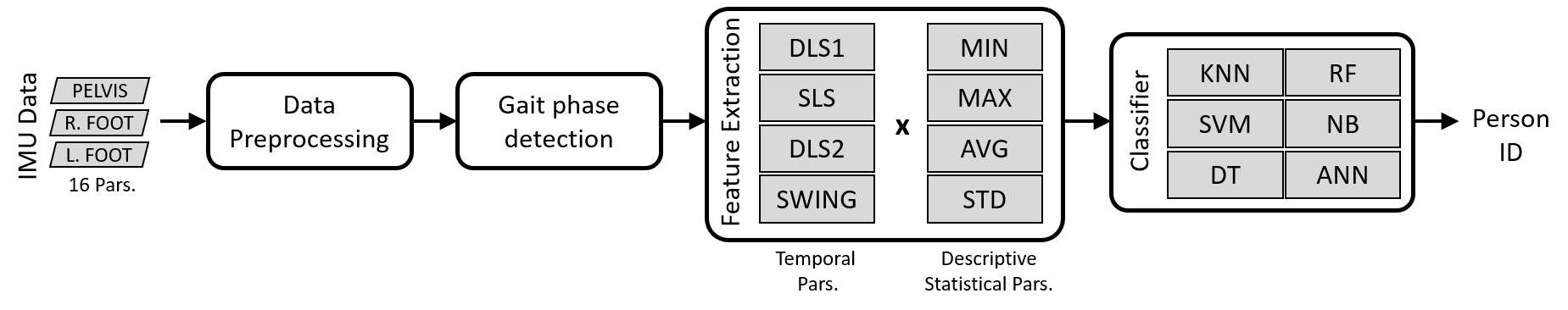}
  \caption{Overview of proposed gait identification methodology}
  \label{Framework}
\end{figure*}

Person identification is an essential prerequisite for initiating any personalized service in a smart living space. The majority of well-known biometric techniques in practice today have their limitations. First, most of these just provide authentication at a specific instant or location that is not suitable for continuous monitoring or tracking. Physiological techniques such as fingerprint, iris, or vein scans all require physical contact with scanning devices. Another important aspect is the vulnerability of appearance-based biometric systems, which could make them more prone to imitation. Appearance-based techniques like face recognition can have problems in the presence of occlusion. A typical example of such a problem is the mass usage of face masks as a result of the recent pandemic outbreak. This has also raised a new level of surveillance and security concerns. In such a scenario, it is highly desirable to identify humans unobtrusively from a distance regardless of changes in their appearance. All these concerns have directed new attention towards behavioral biometrics, especially gait-based human identification, which could overcome these problems.

The biological characteristics of a person's body, including age, gender, height, weight, and health condition, result in a unique pattern of gait motion which can be used as a human biometric identification \cite{nixon2010human,nixon1999automatic}. Apart from being contactless, gait-based systems are hard to fake. The two main methods used for gait monitoring are the wearable sensors and image-based methods \cite{muro2014gait}, however, the former is preferred for its continuous and accurate measurement without any dependence on view angle pose and visibility. Utilizing minimum time and resources to accurately accomplish the recognition task is a concern for all methods. One of the things that makes gait recognition (and most of the other behavioral biometrics) difficult is that it requires a time sequence of information about the position and movement of the person's body joints. And the problem arises when this data cannot be retrieved for the required period continuously. A common example of this would be video-based gait recognition systems where one of the two scenarios can occur: (1) The subject's complete body joints are visible in the video, however, only for a fraction of the complete gait cycle duration, (2) The subject is visible for at least a complete gait cycle duration, however, not all of his body joints are clearly visible. These issues may result from self-occlusion, occlusions from the external environment, or other subjects. Therefore, to alleviate this concern, in this work we investigate and identify the minimum time frame in the whole cycle that contains key features adequate for accurate recognition. In addition to locating the period of interest, we also find which locations on the body hold more valuable information. Apart from resolving the problems mentioned earlier where limited information is available, using the sensor data for only some fraction of the whole gate cycle, and reducing the number of sensors to be monitored will also result in faster computation times. Furthermore, for wearables, this can yield a significant reduction in energy consumption, which is one of the main challenges they face \cite{williamson2015data}.

Inertial measurement unit (IMU) sensors are known for their accuracy in the analysis of complex motion. Their lightweight and small size make them ideal for monitoring human gait motion at different joint locations. They are embedded with an accelerometer and gyroscope, which can be used to precisely analyze the translational and angular dynamics of the gait in the 3-axes. IMU sensors have been widely used for gait-related tasks. These include gait phase detection \cite{abhayasinghe2014human,seel2014online,mannini2012gait}, prediction of biological variables \cite{riaz2015one, hu2018machine}, detection of abnormalities \cite{caramia2018imu, mannini2016machine}, activity recognition \cite{bulling2014tutorial, altun2010comparative, frank2010bayesian} and person identification \cite{sprager2015inertial, huang2007gait, trung2012performance, zhang2012individual}.

Regarding the use of a minimum number of sensors, a lot of work has been done by using a single inertial sensor located on the pelvis or foot mostly for activity recognition \cite{ghobadi2017robust, abhayasinghe2014human, song2017ambulatory}, estimation of temporal events and parameters \cite{trojaniello2014accuracy, panebianco2018analysis, storm2016gait}. For person identification, a single sensor on the pelvis \cite{sama2013gait, nickel2013classifying}, foot \cite{huang2007gait} and ankles \cite{sun2012curve} have been illustrated. The effect of sensor location on identification performance has been discussed in \cite{dehzangi2017imu} and more recently in \cite{icinco20}, however, the number of subjects used is not enough in both ( 10 and 20 respectively) and they use complete stride for classification instead of a fraction. To the best of the author's knowledge, the comparison of different gait phases in terms of person identification accuracy has not been clearly investigated in the literature.

In this paper, we investigate the monitoring locations and gait phases critical for an acceptable person identification performance using minimal IMU sensors. The main contributions of this study can be summarized as follows:
\begin{enumerate}
  \item We propose an identification methodology based on temporal parameters of the gait combined with descriptive statistics of different gait phases as the features. 
  \item Three sensor locations and their combinations have been investigated to find the most suitable sensor location.
  \item By using only a fraction of the complete gait cycle, we investigate the most valuable gait phase that gives a satisfactory gait identification accuracy.
  \item We use and compare 6 well-known machine learning techniques using the proposed input features to evaluate the identification performance.
\end{enumerate}
The proposed methodology used in this paper is depicted in Fig. \ref{Framework}. First, the IMU data is preprocessed after collection. Next, the data is segmented to extract individual gait phases. Temporal and descriptive statistics parameters are then extracted and used as input to classifiers for identification. The rest of the paper is organized as follows. The data collection, preprocessing, gait cycle overview, feature extraction, and identification are presented in Section II. Results are analyzed and discussed in Section III and finally, Section IV concludes the work.

\section{Methodology}

The following subsections describe the proposed methodology steps. 

\subsection{Data collection}

Data were collected from a total of 60 individuals, with ages between 20s to 60s. A commercial IMU-based motion capture system (Xsens MVN, Enschede,
Overijssel, Netherland) was utilized. A total of 3 IMU sensors were attached to each subject, one on the pelvis and one on each foot. The subjects were instructed to walk at their preferred speed and an average of 9 strides were collected from each subject. For each sensor, accelerometer, and gyroscope data was collected at a sampling rate of 1000S/s.

\begin{figure*}[thpb]
      \centering
      %\framebox{\parbox{3in}{We suggest that you use a text box to insert a graphic (which is ideally a 300 dpi TIFF or EPS file, with all fonts embedded) because, in an document, this method is somewhat more stable than directly inserting a picture.}
      \includegraphics[width=0.85\linewidth]{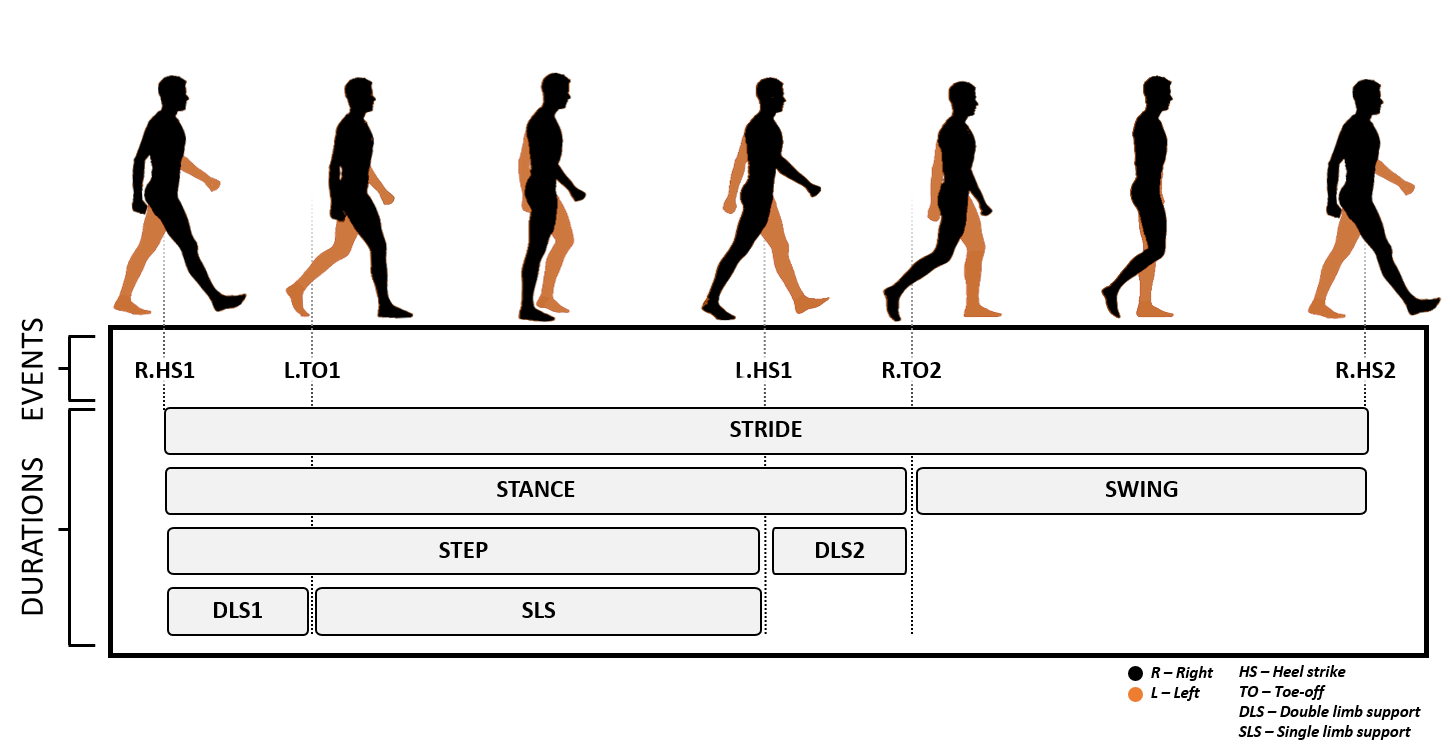}

      \caption{Phases and subphases of human gait cycle}
      \label{GaitCycle}
   \end{figure*}

\subsection{Data Preprocessing}

 Noise in the form of high-frequency spikes in the raw data could hinder the detection of gait phases. Therefore, a 3 order, zero-lag Butterworth band-pass (0.5-15Hz) filter was used to filter high-frequency noise from the raw data. 
 To overcome the location-based errors in the accelerometer output, all accelerometer data were scaled to the range 0 to 1 using the MinMax scaling which preserves the shape of the original distribution. This makes the raw data more comparable between subjects.
 Furthermore, the sensors at the feet are prone to errors in orientation. This sometimes results in intermixing of acceleration and angular velocity measured along the X and Y axes resulting in data from different subjects difficult to compare. To alleviate this error, we compute new combined magnitude measurements along the two axes for the feet sensors as follows
 \begin{equation} \label{eq:1} Mag(t) = R_{xy} = \sqrt{{R_x(t)}^2 +{R_y(t)}^2}\end{equation}
 %$R_{XY} = sqrt{{R_X}^2 +{R_Y}^2}$

\subsection{Overview of human gait cycle}
To understand the feature extraction and interpret the results from different phases of the human gait cycle, it is important to give a brief overview of these phases. The normal human gait cycle starts as the heel of the reference foot contacts the ground and ends when it contacts again as shown in Fig. \ref{GaitCycle}. The gait cycle is also referred to as a stride. The stride can be broken into mainly the stance and swing phases \cite{o2019physical}. In the stance phase, the reference foot is in contact with the ground surface, while in the swing phase it remains off the ground. The stance and swing phases normally constitute about 60\% and 40\% of the total gait cycle respectively \cite{perry1992gait}. The stance phase can be further divided into three subphases: initial double-limb support (DLS1), single-limb support (SLS), and terminal double-limb support (DLS2). DLS1 is the duration at the start of the stance phase which starts as the reference foot heel strikes the ground, called the heel-strike (HS) event. In this subphase, the body is shifting its weight from reference foot in the front to rear foot and both feet are in contact with the ground. Finally, as the reference foot reaches a flat position on the ground, the contralateral foot rises from the ground, an event known as the toe-off (TO). This marks the start of SLS as only the reference foot is in contact with the ground. The combined duration of DLS1 and SLS is referred to as a step. The SLS ends as DLS2 starts when the contralateral foot is stretched forward and strikes the ground, called the heel-strike event for that foot. For the duration of DLS2, both feet remain in contact with the ground until the toe-off event occurs for the reference foot marking the start of the swing phase of the reference foot. In this phase, the reference foot is swung from the back and stretched to the front. The swing phase ends with a heel strike event for the reference foot, marking the end of the current gait cycle.

\begin{figure}[b]
      \centering
      \includegraphics[width=\linewidth]{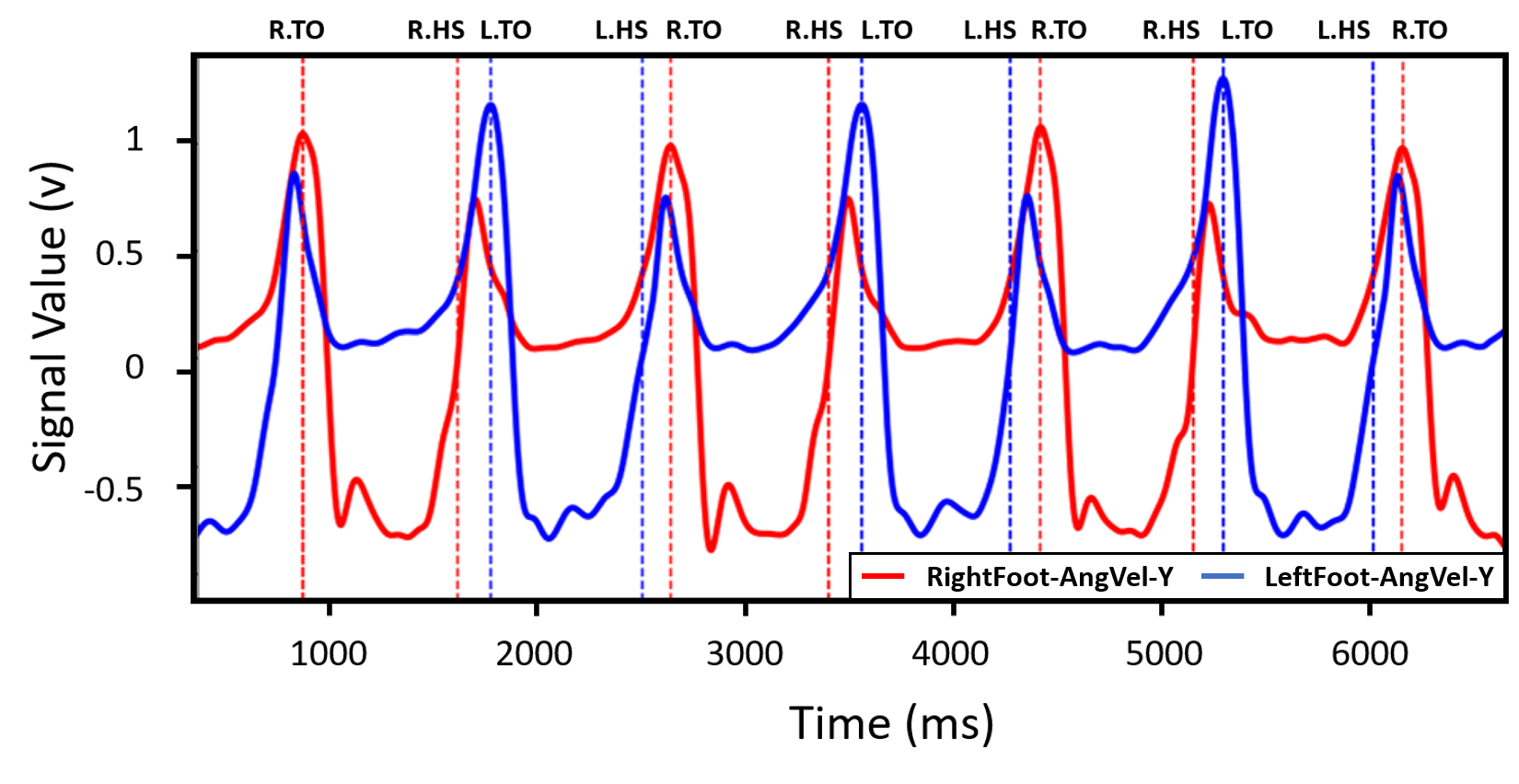}
      \caption{An example of gait events detected through the Angular Velocity Y signal}
      \label{GyroscopeSignal}
   \end{figure}

\subsection{Feature extraction} 

To extract the features, the gait cycle events heel-strike and toe-off were first detected using the angular velocity measured along the Y-axis at the two foot-sensors as shown in Fig. \ref{GyroscopeSignal}. The signal consists of alternating peaks of low and high amplitude. Here the toe-off was detected at the high-amplitude peaks and the heel-strike at the point where the signal crosses the zero line before the low-amplitude peak \cite{jasiewicz2006gait}.
Detection of these events enables computation of all gait-phase intervals. From here we first extract temporal parameters DLS1, SLS, DLS2, and swing for each adjacent right and left foot strides. Next, to capture the signal dynamics during these phase intervals, we propose using descriptive statistics parameters as features. Compared to complex frequency domain-based or explicitly defined time-domain features, descriptive statistics provide simple yet powerful means of quantitatively summarizing the signal dynamics in the selected intervals. We compute 4 basic descriptive statistics parameters; minimum (MIN), maximum (MAX), mean (AVG), and standard deviation (STD) of each sensor signal for each phase. By combining the temporal parameters with descriptive statistics parameters we are essentially extracting both the temporal and kinematic variations of the gait which enable the classification networks to distinguish and recognize gaits of different individuals.
Computing the 4 descriptive statistics parameters for each of the 6 signals from the pelvis sensor and 4 from each foot (as we measure magnitude parameters from X and Y axes), we get a total of 24 descriptive statistics parameters. Combining these with the 8 temporal parameters, we extract 32 features in total for each stride from a subject. These 32 features are extracted from an individual gait phase. The number of features multiplies as we combined more than a single phase for identification.
%Computing the 4 descriptive statistics parameters for each of the 6 signals from the pelvis sensor and 4 from each foot (as we measure magnitude parameters from X and Y axes), we get 56 descriptive statistics parameters for the case using all sensors. Adding the temporal parameter, i.e. the duration for selected phase, we extract 57 features in total. The number of features multiplies as we combined more than a single phase for identification.

\subsection{Training and Identification} The feature data were scaled using the MinMax scaling before training. Since our data is composed of a different number of strides for each subject, therefore we use the stratified k-fold cross-validation procedure for training and testing different classification algorithms with an 80/20 training-test split. Six different machine learning techniques including k-Nearest Neighbors (KNN), Support Vector Machines (SVM), Random Forest (RF), Decision Tree (DT), Naive Bayes (NB), and Artificial Neural Networks (ANN) were employed and compared. 
For each classifier, a range of different parameter values was tested and evaluated on the dataset to find their optimal values that result in the highest classification accuracy. For KNN, the optimal k value was found to be 1 through a bootstrap procedure. In SVM, the C parameter was set to 1 and linear type kernel was used. For RF, the optimal values for the number of trees and the maximum depth of the tree were found to be 100 and 30 respectively. The optimal value for minimum samples to split for DT was found to be 2. For the ANN, the optimal set of parameters was identified using the grid search technique. A two hidden layer structure with 500 neurons in each layer was found to be the best configuration. We used the Adam optimizer with a constant learning rate of 0.001. The activation function selected in the hidden and output layers is the hyperbolic tan function, and softmax function respectively.

% \usepackage{multirow}
\begin{table}[h]
\caption{Average accuracy results from 5-fold cross-validation}
\begin{tabular}{ll|l|l|l|l|l|l|}
\cline{3-8}
                                                     &       & KNN   & SVM   & DT    & RF    & NB    & ANN   \\ \toprule
\multicolumn{1}{|l|}{\multirow{7}{*}{\rotatebox[origin=c]{90}{FOOT}}}          & DLS1   & 0.766 & 0.600 & 0.577 & \textbf{0.796} & 0.764 & 0.768 \\ 
\multicolumn{1}{|l|}{}                               & SLS    & 0.834 & 0.572 & 0.564 & \textbf{0.881} & 0.826 & 0.866 \\ 
\multicolumn{1}{|l|}{}                               & DLS2   & 0.825 & 0.694 & 0.574 & 0.821 & 0.806 & \textbf{0.828} \\ 
\multicolumn{1}{|l|}{}                               & SWING  & 0.825 & 0.600 & 0.609 & 0.885 & 0.779 & \textbf{0.862} \\
\multicolumn{1}{|l|}{}                               & STEP   & 0.877 & 0.866 & 0.613 & 0.921 & 0.853 & \textbf{0.923} \\
\multicolumn{1}{|l|}{}                               & STANCE & 0.953 & \textbf{0.979} & 0.589 & 0.953 & 0.891 & 0.964 \\
\multicolumn{1}{|l|}{}                               & STRIDE & 0.964 & \textbf{0.987} & 0.655 & 0.975 & 0.921 & 0.977 \\ \midrule
\multicolumn{1}{|l|}{\multirow{7}{*}{\rotatebox[origin=c]{90}{PELVIS}}}        & DLS1   & 0.815 & 0.798 & 0.543 & 0.862 & 0.808 & \textbf{0.874} \\ 
\multicolumn{1}{|l|}{}                               & SLS    & 0.933 & 0.861 & 0.613 & 0.932 & 0.878 & \textbf{0.942} \\ 
\multicolumn{1}{|l|}{}                               & DLS2   & 0.821 & 0.774 & 0.557 & \textbf{0.858} & 0.825 & \textbf{0.858} \\ 
\multicolumn{1}{|l|}{}                               & SWING  & 0.926 & 0.885 & 0.604 & 0.926 & 0.889 & \textbf{0.955} \\ 
\multicolumn{1}{|l|}{}                               & STEP   & 0.955 & 0.970 & 0.604 & 0.966 & 0.915 & \textbf{0.981} \\ 
\multicolumn{1}{|l|}{}                               & STANCE & 0.981 & \textbf{0.989} & 0.577 & 0.987 & 0.943 & 0.981 \\ 
\multicolumn{1}{|l|}{}                               & STRIDE & 0.992 & \textbf{0.994} & 0.662 & 0.991 & 0.943 & 0.991 \\ \midrule
\multicolumn{1}{|l|}{\multirow{7}{*}{\rotatebox[origin=c]{90}{FOOT + PELVIS}}} & DLS1   & 0.909 & \textbf{0.960} & 0.609 & 0.925 & 0.874 & 0.957 \\ 
\multicolumn{1}{|l|}{}                               & SLS    & 0.979 & 0.979 & 0.660 & 0.975 & 0.932 & \textbf{0.991} \\ 
\multicolumn{1}{|l|}{}                               & DLS2   & 0.972 & \textbf{0.981} & 0.609 & 0.945 & 0.909 & 0.958 \\ 
\multicolumn{1}{|l|}{}                               & SWING  & \textbf{0.985} & 0.975 & 0.685 & 0.974 & 0.913 & 0.968 \\ 
\multicolumn{1}{|l|}{}                               & STEP   & 0.987 & 0.996 & 0.649 & 0.970 & 0.938 & \textbf{0.998} \\ 
\multicolumn{1}{|l|}{}                               & STANCE & 0.994 & \textbf{0.998} & 0.651 & 0.994 & 0.947 & 0.996 \\ 
\multicolumn{1}{|l|}{}                               & STRIDE & \textbf{1.000} & \textbf{1.000} & 0.657 & 0.992 & 0.960 & \textbf{1.000} \\ \bottomrule
\end{tabular}
\label{table}
\end{table}

\section{RESULTS AND DISCUSSION} The mean accuracies obtained from the 5-fold cross-validation (CV) method for different configurations are given in Table \ref{table}. The highest accuracy obtained for each interval (row) is highlighted in bold. From these results, it can be observed that ANN and SVM both have the best performance overall while DT yields the least accuracy. Another important observation that can be made about SVM and ANN from the table is that the SVM yields higher accuracy when a high number of features are provided. This is evident from the results as the SVM results in high accuracy for either when a longer portion of the gait cycle was used (STANCE and STRIDE) or when both pelvis and foot-sensors were monitored. In contrast, the ANN is more robust to the number of input data features as it also yields high accuracy for both single gait phases and single monitoring sites. Therefore, it can be concluded that ANN is overall the best machine learning algorithm for gait identification. The maximum of 100\% identification accuracy was achieved when a complete stride (gait-cycle) was monitored using both pelvis and foot sensors by employing the KNN, SVM, or ANN.

\begin{figure}[b]
      \centering
      \includegraphics[width=0.88\columnwidth]{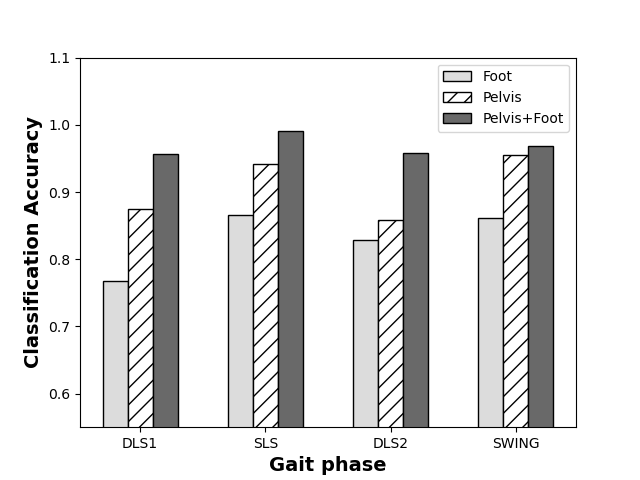}
      \caption{Individual gait phase comparison through ANN results}
      \label{phaseComparisonCombinedAvg}
   \end{figure}

From the results, we can compare the identification performance of individual gait phases of the gait cycle. We illustrate this by charting the ANN results in Fig. \ref{phaseComparisonCombinedAvg}. It is evident from these results that SLS and SWING phases have higher accuracy compared to DLS1 and DLS2. This result was expected because in the two double limb support phases DLS1 and DLS2, both feet are in contact with the ground and have fewer signal dynamics compared to the SLS and SWING phases where one foot swings from back to the front. Hence, it can be stated that the SLS or SWING phase are more critical intervals for monitoring in the overall gait cycle for gait identification. Now, to determine the possibility of using only a single gait phase for achieving high identification accuracy, the results show that high identification accuracy is possible when monitoring a single phase in the whole gait cycle even through using a single sensor. As shown in Fig. \ref{phaseComparisonCombinedAvg}, using the pelvis sensor in the SWING phase can yield a high accuracy of 95.5\% which is obtained from the ANN. Similarly, the SLS phase gives the best results with an accuracy of 94.2\% when monitoring the pelvis only. These are the two most efficient configurations which use a single sensor for only a fraction (2/5th) of the gait cycle and are still able to achieve a high recognition accuracy.  If we include both the feet and the pelvis these accuracies can be increased to 99.1\% and 96.8\% respectively.

\begin{figure}[t]
      \centering
      \includegraphics[width=0.88\columnwidth]{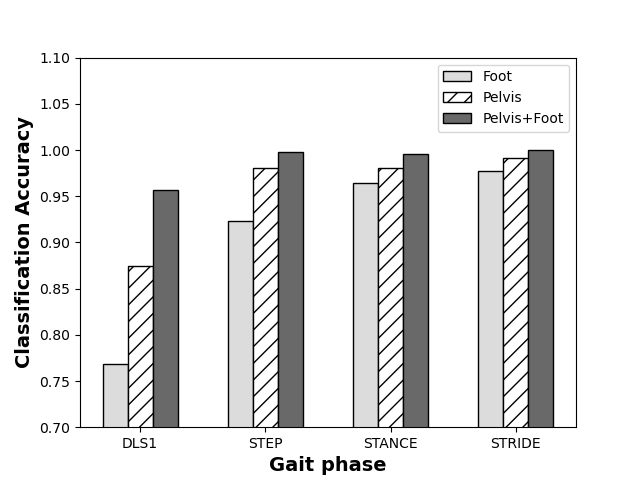}
      \caption{Increase in accuracy with increasing fraction of gait cycle used for classification}
      \label{phaseComparisonCombined}
   \end{figure}

It is also important how the identification performance improves as we start from the first gait phase (DLS1) and start combining the phases sequentially up to the complete gait cycle. As the classifier gets a higher fraction of the complete gait cycle, its performance improves until reaching the maximum when its provided with the complete gait cycle. This can be observed in Fig. \ref{phaseComparisonCombined} where the ANN results are charted separately.

\begin{figure}[t]
      \centering
      \includegraphics[width=0.88\columnwidth]{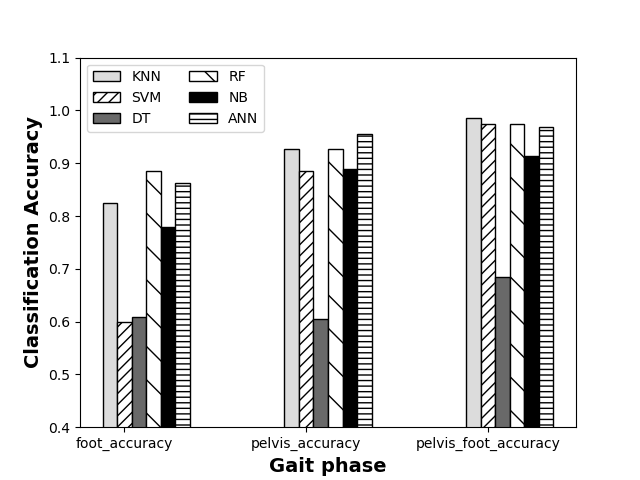}
      \caption{A comparison of different monitoring site configurations}
      \label{sensorComparison}
   \end{figure}

The performance of the three configurations of monitoring locations; foot, pelvis, and both combined for the SWING phase is illustrated separately in Fig. \ref{sensorComparison} for each of the techniques. It can be observed that the overall pelvis is a better monitoring site for gait identification compared to the feet. This is because the pelvis motion accounts for three of the six gait determinants, namely pelvic rotation, obliquity, and lateral displacement of the pelvis. Thus pelvis sensor can observe dynamics from all kinds of body movements in the three axes. Furthermore, the pelvis is less prone to sensor location and orientation errors compared to the sensors in the feet.

\begin{figure}[b]
      \centering
      \includegraphics[width=\linewidth]{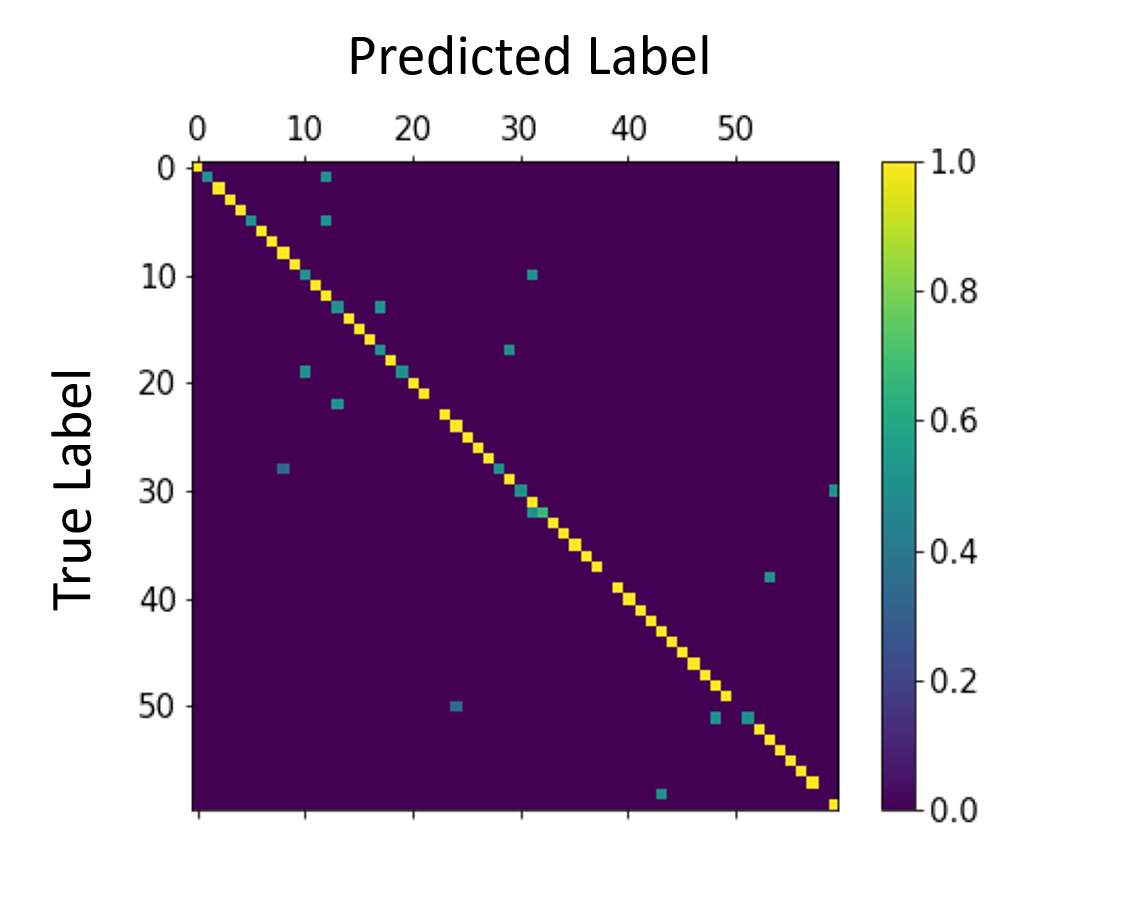}
      \caption{Normalized confusion matrix for ANN as an example when monitoring pelvis for the DLS1 phase}
      \label{confMatrix}
   \end{figure}

In Fig. \ref{confMatrix} the confusion matrix for identification through ANN has been presented as an example when only the pelvis was used as the monitoring site for the duration of the DLS1 phase. This case has been selected to observe the predictions when the accuracy is not 100\%. As shown, most of the individuals were identified correctly. For the incorrect predictions, not all steps from an individual were incorrectly predicted. It is interesting that for the four individuals with 0\% identification accuracy (labeled 22, 38, 50, and 58) the number of extracted strides was 4, 5, 8, and 8 respectively. This shows that the incorrect predictions are not made because of the lower number of samples. These failures could be a result of other factors in the features extracted for these individuals. Nonetheless, it must be noted that all individuals in the dataset including these failed ones were identified correctly using both pelvis and foot sensors for the duration of the complete gait cycle. 

\section{CONCLUSIONS}

An IMU-based gait identification methodology was proposed, using the temporal parameters of the gait combined with descriptive statistics parameters as gait modeling features. We investigated the possibility of using a single gait phase for identification with a single sensor monitoring site. For the most efficient configuration, using only a single sensor with monitoring of only a single gait phase, a high accuracy of 95.5\% was achieved through the ANN (with pelvis sensor for SWING phase). A 100\% identification accuracy was achieved when all pelvis and feet sensors were used for the whole gait cycle through KNN, SVM, and ANN. It can be concluded that a reasonably accurate human identification even with a single sensor can be achieved provided that the suitable intervals are monitored correctly. These findings can be useful for a more efficient and robust gait-based human identification through any sensing methodologies in the future.       

\addtolength{\textheight}{-11cm}   % This command serves to balance the column lengths

%%%%%%%%%%%%%%%%%%%%%%%%%%%%%%%%%%%%%%%%%%%%%%%%%%%%%%%%%%%%%%%%%%%%%%%%%%%%%%%%

%%%%%%%%%%%%%%%%%%%%%%%%%%%%%%%%%%%%%%%%%%%%%%%%%%%%%%%%%%%%%%%%%%%%%%%%%%%%%%%%

%%%%%%%%%%%%%%%%%%%%%%%%%%%%%%%%%%%%%%%%%%%%%%%%%%%%%%%%%%%%%%%%%%%%%%%%%%%%%%%%
\bibliographystyle{IEEEtran}
\bibliography{IEEEabrv,IEEEexample}
\end{document}